\def\year{2021}\relax
\documentclass[letterpaper]{article} 
\usepackage{aaai21}  
\usepackage{times}  
\usepackage{helvet} 
\usepackage{courier}  
\usepackage[hyphens]{url}  
\usepackage{graphicx} 
\usepackage{natbib}  
\usepackage{caption} 
\newcommand{\etal}{\textit{et al}. }

\usepackage{graphicx} 
\usepackage{amssymb}
\usepackage{threeparttable}
\usepackage{multicol}
\usepackage{multirow}
\usepackage{colortbl}
\usepackage{bm}
\usepackage{enumitem}
\usepackage{algorithm}
\usepackage{algpseudocode}
\usepackage{blindtext}
\usepackage{amsmath}
\usepackage{wrapfig}
\usepackage{capt-of}
\usepackage{tabularx}
\usepackage[export]{adjustbox}
\usepackage{booktabs}
\usepackage{pifont}
\usepackage{lipsum}
\usepackage{color}
\usepackage{url}

\DeclareMathOperator*{\argmin}{arg\,min}

\definecolor{mygreen}{rgb}{0.2,0.5,0.2}

\title{Up to 100$\times$ Faster Data-free Knowledge Distillation}
\author {
    Gongfan Fang\textsuperscript{\rm 1,3}\thanks{Equal contributions},
    Kanya Mo\textsuperscript{\rm 1}$^{*}$,
    Xinchao Wang\textsuperscript{\rm 2},
    Jie Song\textsuperscript{\rm 1}\\
    Shitao Bei\textsuperscript{\rm 1},
    Haofei Zhang\textsuperscript{\rm 1},
    Mingli Song\textsuperscript{\rm 1}\thanks{Corresponding author} \\
}
\affiliations {
    \textsuperscript{\rm 1}Zhejiang University \\
    \textsuperscript{\rm 2}National University of Singapore \\
    \textsuperscript{\rm 3}Alibaba-Zhejiang University Joint Research Institute of Frontier Technologies \\
    \{fgf, sjie, best, haofeizhang, brooksong\}@zju.edu.cn\\
    Kanya.18@intl.zju.edu.cn, xinchao@nus.edu.sg
}

\begin{document}

\maketitle

\begin{abstract}
    Data-free knowledge distillation (DFKD) has recently been attracting increasing attention from research communities, attributed to its capability to compress a model only using synthetic data. Despite the encouraging results achieved, state-of-the-art DFKD methods still suffer from the inefficiency of data synthesis, making the data-free training process extremely time-consuming and thus inapplicable for large-scale tasks. In this work, we introduce an efficacious scheme, termed as FastDFKD, that allows us to accelerate DFKD by a factor of orders of magnitude. At the heart of our approach is a novel strategy to reuse the shared common features in training data so as to synthesize different data instances. Unlike prior methods that optimize a set of data independently, we propose to learn a meta-synthesizer that seeks common features as the initialization for the fast data synthesis. As a result, FastDFKD achieves data synthesis within only a few steps,  significantly enhancing the efficiency of data-free training. Experiments over CIFAR, NYUv2, and ImageNet demonstrate that the proposed FastDFKD achieves 10$\times$ and even 100$\times$ acceleration while preserving  performances on par with state of the art. Code is available at \url{https://github.com/zju-vipa/Fast-Datafree}.
\end{abstract}

\section{Introduction}
Knowledge distillation~(KD) has recently emerged as a popular paradigm
to reuse pre-trained models that are nowadays prevalent online.
KD aims  to train a  compact  student model for efficient inference by imitating the behavior of the pre-trained teacher~\cite{hinton2015distilling,yang2020distilling,fang2021mosaicking}.
The conventional setup of KD
requires possessing the original training data as input so as to train the student. 
Unfortunately, due to confidential or copyright reasons, 
in many cases the original data cannot be released 
and only the pre-trained models are available to users~\cite{kolesnikov2020big,shen2019amalgamating,ye2019student},
which, in turn, imposes a major obstacle towards applying  KD 
for a broader domain.


\begin{figure}[t]
   \begin{center}
      \includegraphics[width=\linewidth]{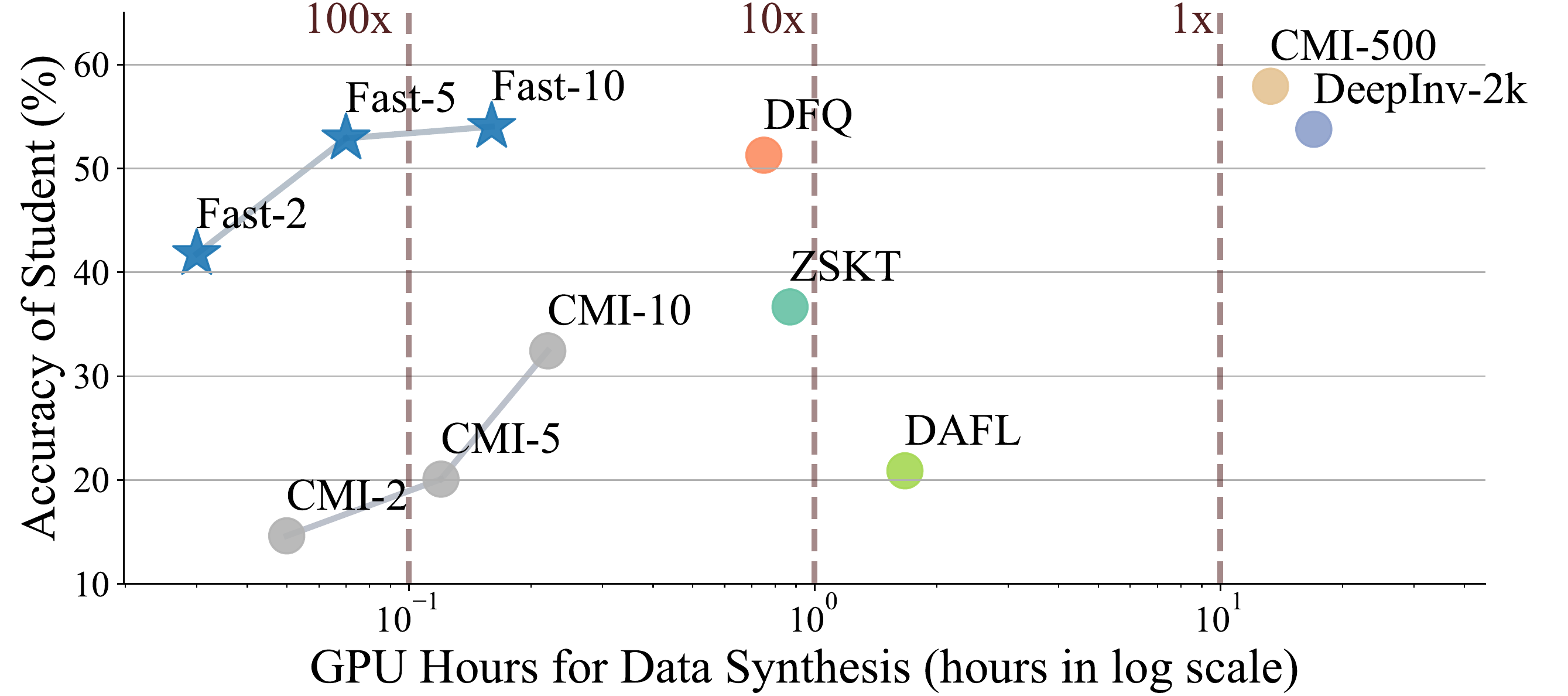}
   \end{center} 
      \caption{Accuracy (\%) of student models v.s. GPU hours of data synthesis on CIFAR-100 dataset. Our method, denoted as ``Fast'', achieves 10$\times$ to even 100$\times$ acceleration and  performance on par with existing methods.} \label{fig:intro}
\end{figure}

To remedy this issue,
data-free knowledge distillation (DFKD) approaches 
have been  proposed by assuming 
that no access to training data is available at all~\cite{lopes2017data}.
Due to the much relaxed constraint on training data,
DFKD has been receiving increasing attention from research communities including computer vision~\cite{chen2019data}, natural language processing~\cite{ma2020adversarial}, and graph learning~\cite{deng2021graph}. 
Typically, DKFD follows a distilling-by-generating paradigm, where a synthetic dataset is often crafted for training by ``inverting'' the pre-trained teacher~\cite{yin2019dreaming}. 
To learn a comparable student model, the synthetic set should contain sufficient samples to
enable a comprehensive knowledge transfer from teachers.
Consequentially, this poses a significant challenge to DFKD, 
since synthesizing a large-scale dataset is inevitably time-consuming, especially for sophisticated tasks like ImageNet recognition~\cite{yin2019dreaming} and COCO detection~\cite{chawla2021data}. 


In this paper, we introduce a novel approach,
termed as FastDFKD,
to accelerate 
data synthesis process so as to make data-free knowledge distillation 
more applicable for large-scale tasks.
Our motivation stems from the fact that, 
data instances from the same domain usually share some common features,
and hence such shared features should be explicitly utilized for 
data synthesis. 
For example, the textures of ``fur'' may frequently emerge in an animal dataset,
and thus {can be reused to create different instances}.
Unfortunately, 
existing DFKD approaches
have mainly focused on synthesizing
samples independently
and, in fact,
none of the existing DFKD approaches has explored
taking advantage of feature sharing,
making the DFKD process very cumbersome. 

The proposed FastDFKD approach, on the other hand, expressively
explores the common features between samples for synthesizing.
FastDFKD follows the batch-based strategy for data synthesis~\cite{yin2019dreaming,fang2021contrastive};
yet unlike prior methods that synthesizes different samples independently,
FastDFKD focuses on a ``learning to synthesize'' problem, where an efficient synthesizer is explicitly trained to capture common features for fast adaptation. The advantage of common features sharing is that we do not need to synthesize them repeatedly for each instance, which significantly improves the synthesis efficiency.

Specifically, we develop FastDFKD under a meta-learning framework~\cite{finn2017model}, which aims to learn a meta-generator in the synthesis process. FastDFKD comprises two optimization loops, the outer loop, and the inner loop. The inner loop accounts for the data synthesis process, where a set of samples are created by adapting and re-organizing common features. The outer loop, on the other hand, updates the common features using the results of inner loops for a better meta-initialization. As illustrated in Figure \ref{fig:intro}, such a meta synthesizer dramatically improves the efficiency of data synthesis in data-free distillation while preserving a performance on par with state of the art. As will be demonstrated in our experiments,
FastDFKD is able to achieve 10$\times$ and in some cases even more than 100$\times$ 
speed up compared to the state of the art.

Our contribution is therefore a novel DFKD scheme, 
termed as FastDFKD,
that allows us to significantly accelerate the data-free training
through common feature reusing. 
Experimental results
over the CIFAR, ImageNet, and NYUv2 datasets 
demonstrate that,
FastDFKD yields  performances comparable to the state of the art,
while achieving a speedup
factor of 10 and sometimes even more than 100
over existing approaches. 

\section{Related Works}
\paragraph{Data-Free Knowledge Distillation.}
Data-free knowledge distillation aims to train a compact student model from a pre-trained teacher model without access to original training data. It typically follows a distilling-by-generating paradigm, where a fake dataset will be synthesized and used for student training. In the literature, Lopes \etal proposes the first data-free approach for knowledge distillation, which utilizes statistical information of original training data to reconstruct a synthetic set during knowledge distillation~\cite{lopes2017data}. This seminal work has spawned several works, which has achieved impressive progressive on several tasks including detection~\cite{chawla2021data}, segmentation~\cite{fang2019data}, text classification~\cite{ma2020adversarial}, graph classification~\cite{deng2021graph} and Federated Learning~\cite{zhu2021data}. Despite the impressive progress, a vexing problem remains in DFKD, i.e., the inefficiency of data synthesis, which makes data-free training extraordinarily time-consuming. For example, \cite{luo2020large} trains 1,000 generators to compress an ImageNet-pretrained ResNet-50 and \cite{yin2019dreaming} optimizes a large number of mini-batches for data synthesis. In this work, we focus on this under-studied problem, i.e., the efficiency of DFKD, and proposes the first approach to accelerate data-free training.

\paragraph{Meta Learning.} Meta-learning is a popular framework for few-shot learning~\cite{hospedales2020meta}, which follows a ``learning to learn'' paradigm to find a helpful initialization for target tasks. Among various meta-learning algorithms, MAML is one of the most influential methods owning to its impressive results on several benchmarks ~\cite{finn2017model,nichol2018first}. As an optimization-based method for meta-learning, MAML introduces two optimization loops to handle a set correlated tasks: an inner loop for task learning and an outer loop for training a meta-learner. The inner and outer loops are collaboratively trained to find a meta-initialization that can be adapted to different tasks quickly, where some general knowledge across tasks are captured by the meta-learner~\cite{finn2017model}. Inspired by the ``learning to learn'' paradigm of meta-learning, we develop a fast approach to train a meta-synthesizer for DFKD problems, which can be quickly adapted for fast data synthesis.

\section{Method}
\newcommand{\mathT}[0]{\mathcal{T}}
\newcommand{\mathL}[0]{\mathcal{L}}
\newcommand{\mathG}[0]{\mathcal{G}}

\begin{figure*}[t]
  \begin{center}
      \includegraphics[width=0.8\linewidth]{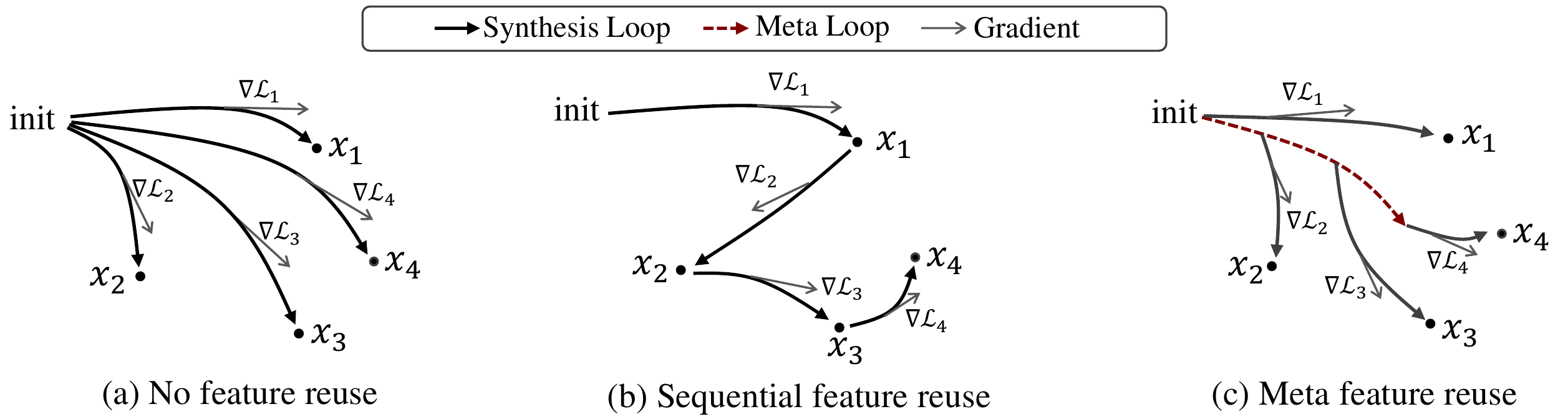}
   \end{center} 
      \caption{Diagram of the proposed meta feature reuse, as well as its difference to other synthesis strategies. (a) Data instances are synthesized independently without feature reuse; (b) Data instances are synthesized sequentially reusing previous results as initialization. (c) The proposed common feature reuse that learns a meta-generator for fast adaptation.}
     \label{fig:framework}
\end{figure*}

\subsection{Problem Setup} 
Given a teacher model $f_t(x; \theta_t)$ pre-trained on a labeled but inaccessible training set $D^t = \sum_i^N\{(x_i, y_i) | x_i\in\mathcal{X}, y_i \in \mathcal{Y}\}$, the goal of data-free knowledge distillation (DFKD) is to craft a synthetic dataset $D = \sum_i^N\{x_i | x_i\in\mathcal{X}\}$ with $N$ samples by inverting the pre-trained model, on which a comparable student model $f_s(x; \theta_s)$ can be trained by imitating the behaviour of the teacher. Typically, The synthesis of $D$ is driven by a pre-trained inversion loss $\mathL: \mathcal{X}\rightarrow\mathbb{R}$, which indicates whether a input sample $x$ comes from the training domain according to some statistical information in the pre-trained teacher model~\cite{yin2019dreaming}. Therefore, the optimization of a single data point $x$ can be formalized as follows:
\begin{equation}
 x^{\ast} = \argmin_{x} \mathL(x)
 \label{eqn:sub_problem}
\end{equation}
To obtain a complete synthetic set $D^{\prime}=\{x_1, x_2, ..., x_N\}$ of size $N$, DFKD repeats the above optimization to construct a set of samples, which leads to a series of optimization problems in the form of Equation \ref{eqn:sub_problem}. Notably, the loss function for different instances $x_i$, denoted as $\mathL_i$, can be different, so that a diverse dataset $D$ can be constructed to carry out comprehensive knowledge from the teacher. To this end, we consider a generalized data synthesis problem for DFKD, which leverages a set of inversion losses $\mathL=\{\mathL_1, \mathL_2, ..., \mathL_N\}$ to craft the synthetic dataset  as follows:
\begin{equation}
 D^{\prime} = \{x^{\ast}_1, x^{\ast}_2, ..., x^{\ast}_N\} = \argmin_{x_1, x_2, ..., x_N} \mathop{\sum}\limits_{i}^N \mathL_{i}(x_i)
 \label{eqn:objective}
\end{equation}
In DFKD, a popular way to solve Equation \ref{eqn:objective} is to optimize different samples directly in a batch-by-batch manner, ~\cite{yin2019dreaming,fang2019data,chawla2021data}. As illustrated in Figure \ref{fig:framework} (a), batch-based approaches synthesize different instances independently and merely take the relation between samples into account. Despite the encouraging results achieved, DFKD  approaches usually suffer from the inefficiency of data synthesis, since crafting a large-scale dataset requires solving a large number of optimization problems in Equation \ref{eqn:sub_problem}, each takes thousands of steps to converge~\cite{yin2019dreaming}. Typically, data from the same domain are likely to share some common features, which can be reused to synthesize different samples. In this work, we present FastDFKD, a novel and efficacious approach to learning common features to accelerate the optimization of Equation \ref{eqn:objective}.

\subsection{Fast Data-Free Knowledge Distillation}

\paragraph{Overview.} At the heart of our proposed approach is the reusing of common features. Our motivation stems from the fact that data from the same domain typically share some reusable patterns, which can be reused to synthesize different instances. The following sections develop a novel definition for common features from a generative perspective and propose FastDFKD to capture common features for fast synthesis through a meta-learning process.

\paragraph{Common Feature.} As the key step towards fast data-free training, a clear definition of the common feature is required to construct an optimizable objective for network training. As illustrated in Figure \ref{fig:framework} (b), a naive reusing strategy would be the sequential feature reusing, where features learned in the previous synthesis are directly used as the initialization to craft new samples. However, such a naive scheme would be problematic since the learned features only come from a single data point, which may not always be reusable for other samples. To address this problem, we develop a more natural definition for common features from a generative perspective. Let's consider a generative network $\mathcal{G}(z; \theta)$ with latent code $z$ and trainable parameters $\theta$, which satisfies that for each sample $x_i \in D^{\prime}$, a latent code $z_i$ can be found to generate $x_i=\mathG(z_i; \theta)$. The generator describes the generation process of different samples $x_i$. To some extent, whether there are common features between a set of samples $D^{\prime}=\{x_1, x_2, ..., x_N\}$ is usually highly correlated with the similarity of different data instances, which means that the generator can implicitly capture the common features if we can find the optimal parameter $\theta$ making the code $z=\{z_1, z_2, ...,z_N\}$ of different samples close in the latent space. Based on this, common features can be learned by solving the following problems:
\begin{equation}
    \min\limits_{z, \theta} \underbrace{\frac{1}{N^2} \sum_{i}\sum_{j} d_z(z_i, z_j)}_{\textit{close in latent space}} +  \underbrace{\frac{1}{N}\sum_{i} d_x(\mathG(z_i, \theta), x_i)}_{\textit{generation}}
    \label{eqn:common_feature}
\end{equation}
where $d_z$ and $d_x$ refers to distance metrics in latent space and input space. The above optimization aims at finding a generation process for dataset $D^{\prime}$, whose $z$ distance in latent space is as small as possible so that, with the learned common features, different samples can be efficiently obtained by navigating in the latent code $z$. However, in data-free settings, the synthetic dataset $D^{\prime}$ is not available until we synthesize it. Thus, We replace the second term defined on $d_x$ of Equation \ref{eqn:common_feature} with the inversion loss $\mathL$ for DFKD, which leads to a data-free objective for common feature learning:
\begin{equation}
    \min\limits_{z, \theta} \underbrace{\frac{1}{N^2} \sum_{i}\sum_{j} d_z(z_i, z_j)}_{\textit{close in latent space}} +  \underbrace{\frac{1}{N}\sum_{i} \mathL_i(\mathG(z_i, \theta))}_{\textit{data-free generation}}
    \label{eqn:datafree_common_feature}
\end{equation}
However, due to the limited capacity of the generative model and the difficulty in GAN training, it is almost intractable to learn a perfect generator to synthesize the full synthetic set $D^{\prime}$ at the same time~\cite{luo2020large}. To remedy this problem, we make some relaxation on Equation \ref{eqn:datafree_common_feature} and does not force the generator $\mathG$ to capture all features for the whole dataset. Instead, we train a generator that allows fast adaptation to different samples within $k$-steps gradient descents, which naturally leads to a meta-learning problem.

\paragraph{Meta Generator.} Equation \ref{eqn:datafree_common_feature} is challenging to optimize because it requires generating the full dataset $D^{\prime}$ with a single generative network, including a lot of non-reusable features. Instead, we can train a generator that only contains common features and synthesize other missing features on the fly for the data synthesis process as illustrated in Figure \ref{fig:framework} (c). Specifically, we relax the objective of common feature learning to train a lightweight generator that can be adapted to synthesize different instances within $k$-step, formalized as a meta-learning problem:
\begin{equation}
    \min\limits_{z, \theta} \frac{1}{N}\sum_{i} \mathL_i(\mathG(\underbrace{U^k_{\mathL_i}(
    \hat{z}, \hat{\theta})}_{\textit{$k$-step adaptation}}))
    \label{eqn:meta_feature}
\end{equation}
where $U^k_{\mathL_i}$ is the inner loop of meta-learning, which refers to a $k$-step optimization initialized from $\hat{\theta}$ and code $\hat{z}$ for the synthesis of $x_i$. The inner loop can be unrolled as follows:
\begin{equation}
    \begin{split}
        z^0_i, \theta^0_i & = \hat{z}, \hat{\theta} \\
        z^k_i &= z^{k-1}_i - \alpha \nabla_{z^{k-1}_i} \mathL_i(\mathcal{G}(z^{k-1}_i;  \theta^{k-1}_i)) \\
        \theta^k_i &= \theta^{k-1}_i - \alpha \nabla_{\theta^{k-1}_i} \mathL_i(\mathcal{G}(z^{k-1}_i; \theta^{k-1}_i))
        \label{eqn:adaptation}
    \end{split}
\end{equation}
Notably, Equation \ref{eqn:meta_feature} plays a similar role as the common feature loss in Equation \ref{eqn:datafree_common_feature}. The inner loop, i.e., the $k$-step adaptation, aims to learn a generator for synthesis by explicitly optimizing the second term of Equation \ref{eqn:datafree_common_feature}. On the other hand, the outer loop tries to make different samples reachable within k-step optimization by implicitly optimizing the first term of Equation \ref{eqn:datafree_common_feature}. 

\paragraph{Optimization.} Optimizing Equation \ref{eqn:adaptation} naturally leads to a meta-learning problem, where a useful initialization $(\hat{z}, \hat{\theta})$ is demanded for fast adaptation. After $k$-step gradient descent, we obtain the a set of new parameter $(z^*_i, \theta^*_i) = U^k_{\mathL_i}(\hat{z}, \hat{\theta}))$ under the guidance of loss function $\mathL_i$, which provides gradient w.r.t $\hat{\theta}$ as follows:
\begin{equation}
\begin{split}
 g_{\hat{\theta}} & = \nabla_{\hat{\theta}} \mathL_{i}(\mathcal{G}(U_{\mathL_i}(\hat{z}; \hat{\theta})))  \\
 & = U^{\prime}_{\mathL_i}(\hat{\theta}) \mathcal{G}^{\prime}(\theta^*)  \mathL^{\prime}_{i}(x^*_i)  \label{eqn:meta_grad}
 \end{split}
\end{equation}
where $\theta^*=U_{\mathL_i}^k(\hat{z}; \hat{\theta})$ refers to the optimization results of $k$-step adaptation using Equation \ref{eqn:adaptation} and $x^*_i = \mathcal{G}(z^*_i; \theta^*_i) = \mathcal{G}(U_{\mathL_i}^k(\hat{z}; \hat{\theta}))$ refers to the synthesis results under the guidance of loss $\mathL_i$. However, note that the $k$-step adaptation in Equation \ref{eqn:adaptation} involves k gradient updates:
\begin{equation}
(z^*, \theta^*) = U^k_{\mathL_i}(\hat{z}; \hat{\theta}) = (\hat{z}, \hat{\theta}) + g_1 + g_2 + ... + g_k
\end{equation}\label{eqn:high_order_grad}
where $g_k$ refers to the gradient computed at the $k$-th step of Equation \ref{eqn:adaptation}, which introduces high-order gradients to the generator training and makes the back-propagation very inefficient. Inspired by prior works in meta learning~\cite{nichol2018first}, we apply a first-order approximation to further accelerate the gradient computing, which treats high-order gradients in \ref{eqn:high_order_grad} as constants and replace the $U^{\prime}_{i}(\hat{\theta})$ with an identity mapping. In this case, the gradient computing in Equation \ref{eqn:meta_grad} only involves first-order gradient and can be simplified as follows:
\begin{equation}
 \nabla_{\hat{\theta}} \mathL_i(\mathcal{G}(U^k_{\mathL_i}(\hat{z}; \hat{\theta}))) = \mathcal{G}^{\prime}(\theta^*)  \mathL^{\prime}_{i}(x^*)  \label{eqn:first_order_meta_grad}
\end{equation}
The first-order approximation directly uses the gradient computed on the adapted generator in the inner loops to update the meta generator. Further, a more efficient gradient approximation, known as reptile~\cite{nichol2018first}, can be achieved by approximating the gradient in Equation \ref{eqn:first_order_meta_grad} with the parameter difference between the adapted generator and meta generator, which further simplifies Equation \ref{eqn:first_order_meta_grad} as:
\begin{equation}
 \nabla_{\hat{\theta}} \mathL_i(\mathcal{G}(U^k_{\mathL_i}(\hat{z}; \hat{\theta}))) = \hat{\theta}_i - \theta^{\ast}_i
 \label{eqn:reptile_meta_grad}
\end{equation}

In conclusion, the optimization of meta generator can be presented as follows:
\begin{equation}
\begin{split}
        \hat{\theta} & = \hat{\theta} - \eta \sum_i \nabla_{\hat{\theta}} \mathL_i(\mathcal{G}(U^k_{\mathL_i}(\hat{z}; \hat{\theta}))) \\
        \hat{z}  & = \hat{z} - \eta \sum_i \nabla_{\hat{z}}  \mathL_i(\mathcal{G}(U^k_{\mathL_i}(\hat{z}; \hat{\theta})))
\end{split}
\label{eqn:meta_update}
\end{equation}

\algdef{SE}[SUBALG]{Indent}{EndIndent}{}{\algorithmicend\ }%
\algtext*{Indent}
\algtext*{EndIndent}
\definecolor{gray}{rgb}{0.5,0.5,0.5}
\algnewcommand{\LineComment}[1]{\State \textcolor{gray}{\(//\) #1}}

\begin{algorithm}[t]
	\caption{FastDFKD}
	\label{alg:alg}
  \begin{flushleft}
    \textbf{Input:} Pretrained teacher $\mathcal{F}_t(x; \theta_t)$, student $\mathcal{F}_s(x; \theta_s)$. \\
    \textbf{Output:} An optimized student $\mathcal{F}_s(x; \theta_s)$ 
  \end{flushleft}
	\begin{algorithmic}[1]
    \State Randomly initialize a generator $\mathcal{G}(\hat{z};\hat{\theta})$
    \State Initialize an empty dataset $D^{\prime} = \{\}$ 
    \For{each synthesis loss $\mathL_i$}:
        \LineComment{1. $k$-step adaptation for synthesis (Eq. \ref{eqn:adaptation})}
        \State Periodically re-initialize $\hat{z}$ for diversity
        \State Reuse meta feature $z, \theta \leftarrow \hat{z}, \hat{\theta}$
        
        \For{k steps}:
            \State Update $(z, \theta) \leftarrow (z, \theta) - \alpha \nabla_{(z, \theta)} \mathL_i( \mathcal{G}(z; \theta) )$ 

        \EndFor
        \State Generate the synthetic data $x^*=\mathcal{G}(z; \theta)$ 
        \State Update the synthetic set $D^{\prime}$ = $D^{\prime} \cup \{x^*\}$
        
        \\
        \LineComment{2. Meta update (Eq. \ref{eqn:reptile_meta_grad})} 
        \State Update $(\hat{z}, \hat{\theta}) \leftarrow (\hat{z}, \hat{\theta}) - \eta \nabla_{(\hat{z}, \hat{\theta})} \mathL_{i}(\mathcal{G}(U^k_{i}(\hat{z}; \hat{\theta})))$ 
        
        \\
        \LineComment{3. KD update (Eq. \ref{eqn:kd_loss})}
        \For{t steps}:
            \State sample a mini-batch $\mathcal{B}$ from $D^{\prime}$
            
            \State $\theta_s \leftarrow \theta_s - \gamma \sum\limits_{x \in \mathcal{B}} \nabla_{\theta_s} KL(f_t(x)\|f_s(x))$ 
        \EndFor
        
    \EndFor

	\end{algorithmic}
\end{algorithm}

\subsubsection{Method Summary.} The proposed method is summarized in Algorithm \ref{alg:alg}, which consists of three stages: 1) a $k$-step adaptation for data synthesis; 2) a meta-learning step for common feature learning; 3) several KD steps to update the student model by optimizing the KL divergence as follows:
\begin{equation}
\theta_s \leftarrow \theta_s - \gamma \sum\limits_{x \in \mathcal{B}} \nabla_{\theta_s} KL(f_t(x)\|f_s(x)) \label{eqn:kd_loss} 
\end{equation}
where $\mathcal{B}$ is a mini-batch sampled from the synthetic set $D^{\prime}$. The proposed approach allows a small $k$ for data synthesis, which significantly improves the efficiency of DFKD. 
\section{Experiments}

\subsection{Inversion Loss for FastDFKD}
In this section, we provide more details about the inversion loss $\mathL$ used in our method. This work mainly focus on classification and segmentation problems, which have been widely studied in the data-free literature~\cite{yin2019dreaming,fang2019data,chen2019data}. We utilize the prevalent Deep Inversion loss proposed by \cite{yin2019dreaming} as criteria for data synthesis and demonstrate how to accelerate the data synthesis with FastDFKD. Note that Deep Inversion loss consists of three terms: a class confidence loss $\mathL_{cls}$, an adversarial loss $\mathL_{adv}$ and a feature regularization loss $\mathL_{feat}$, summarized as follows:
\begin{equation}
\begin{cases}
\mathL_{cls}(x) = CE(f_t(x), c)\\
\mathL_{adv}(x) = - JSD(f_t(x) / \tau \| f_s(x) / \tau) \\
\mathL_{feat}(x) = \sum_{l} \| \mu^l_{feat} - \mu^l_{bn}\| + \| \sigma^l_{feat} - \sigma^l_{bn} \| 
\end{cases}
 \label{eqn:deepinv}
\end{equation}
where both $\mathL_{cls}(x)$ and $\mathL_{adv}(x)$ provide dynamic learning targets for data synthesis and thus can be directly used to construct different synthesis tasks for meta-learning. For example, choosing different pseudo labels can lead to various target categories. However, the loss for feature regularization is unchanged during synthesis since the batch normalization layer only encodes the global statistical information on the whole dataset. To fully leverage the power of meta-learning, which requires a set of different but related losses $\mathL$, we modify the Deep Inversion loss by decomposing the feature regularization loss. Note that the feature regularization $\mathL_{feat}$ aims to solve a distribution matching problem, where the synthetic distribution is will be aligned with the mean and variance $(\mu_{bn}, \sigma^2_{bn})$ stored in BN layers.
The BN statistics is estimated on the whole dataset using the following rules:
\begin{equation}
\begin{split}
\mu_{bn} & = (1 - m) \cdot \mu_{bn} + m \cdot \mu_{feat} \\
\sigma^2_{bn} & = (1 - m) \cdot \sigma^2_{bn} + m \cdot \sigma^2_{feat}
\label{eqn:bn_mmt}
\end{split}
\end{equation}
where $m$ is a momentum parameter for estimating BatchNorm statistics (BNS). Inspired by the momentum estimation of BNS, we propose to construct a dynamic and adaptive feature regularization in a momentum way, too. Specifically, we introduce two accumulative variables to store the mean and variance of already synthesized data and train the inputs to approximate the global BNS as follows:
\begin{equation}
\small
\begin{split}
    \mathL_{feat}(x) = \sum\nolimits_{l} [ & \| (1 - m) \cdot \mu^l_{a} + m \cdot \mu^l_{feat} - \mu^l_{bn}\| + \\ & \| (1 - m) \cdot \sigma^l_{a} + m \cdot \sigma^l_{feat} - \sigma^2_{bn} \| ]
\end{split}
\end{equation}
After synthesis, the accumulative variables $(\mu_{a}, \sigma^2_{a})$ will be updated with Equation \ref{eqn:bn_mmt}, so as to provide a different learning target for meta-learning. 

\subsection{Experimental Settings}
\paragraph{Datasets and models.} We evaluate the proposed method on both classification and semantic segmentation tasks. For image classification, we conduct data-free knowledge distillation on three widely used datasets: CIFAR-10, CIFAR-100~\cite{krizhevsky2009learning} and ImageNet~\cite{deng2009imagenet}. We use the pretrained models from \cite{fang2021contrastive} and follow the same training protocol for comparison, where 50,000 synthetic images are synthesized for distillation. For ImageNet, we use an off-the-shelf ResNet-50 as the teacher and train student models by synthesizing 224$\times$224 images. For semantic segmentation, we use Deeplab models~\cite{chen2017rethinking} trained on NYUv2~\cite{Silberman:ECCV12} dataset for training and evaluation, which contains 1449 densely labeled pairs of aligned RGB images and a 13-class segmentation map. 

\paragraph{Baselines.} Two types of DFKD methods are compared in our experiments: (1) generative methods that train a generative model for synthesis, including DAFL~\cite{chen2019data}, ZSKT~\cite{micaelli2019zero}, DFQ~\cite{choi2020data}, and Generative DFD~\cite{luo2020large} (2) non-generative methods that craft transfer set in a batch-by-batch manner including DeepInv~\cite{yin2019dreaming} and CMI~\cite{fang2021contrastive}.

\paragraph{Evaluation Metrics.} In addition to the standard metrics like classification accuracy and mean IoU for classification and segmentation, we also focus on the efficiency of data synthesis in DFKD, where GPU hours taken by data synthesis is collected and reported. Note that the time cost of student training is omitted since we only focus on the data synthesis process and adopt the vanilla KD~\cite{hinton2015distilling} in all DFKD methods. For fair comparisons, all GPU hours are estimated on a single GPU. 

\newcommand{\std}[1]{\scriptsize $\pm${#1}}
\def\scoreup#1{$(\color{mygreen} \uparrow #1)$}
\def\scoredown#1{$(\color{red} \downarrow #1$)}
\begin{table*}[t]
  \centering
  \small
  \begin{tabular}{c l r r r r r r}
      \toprule
      \bf \multirow{2}{*}{Dataset} & \bf \multirow{2}{*}{Method}  & \bf ResNet-34  & \bf VGG-11 & \bf WRN40-2  & \bf WRN40-2 & \bf WRN40-2 & \bf Average \\
                                  &                     & \bf ResNet-18  & \bf ResNet-18 & \bf WRN16-1  & \bf WRN40-1 & \bf WRN16-2 & \bf Speed Up \\   
    \hline  
      \bf \multirow{11}{*}{\rotatebox[origin=c]{90}{CIFAR-10}} & Teacher  & 95.70  	& 92.25   &	94.87   &	94.87   &	94.87  & - \\ 
      & Student & 95.20   &	95.20   &	91.12   &	93.94   &	93.95   & - \\ 
      & KD  & 95.20   &	95.20   &	95.20   &	95.20   &	95.20   & - \\ 
      & DeepInv$_{2k}$  & 93.26 (42.1h) &	90.36 (20.2h) &	83.04 (16.9h) & 	86.85 (21.9h) &	89.72 (18.2h) & 1.0$\times$ \\ 
      & CMI$_{500}$ &  94.84 (19.0h) &	91.13 (11.6h) &	90.01 (13.3h) &	92.78 (14.1h) &	92.52 (13.6h) & 1.6$\times$ \\ 
      & DAFL & 92.22 (2.73h) &	81.10 (0.73h) &	65.71 (1.73h) &	81.33 (1.53h) &	81.55 (1.60h) & 15.7$\times$ \\ 
      & ZSKT  & 93.32 (1.67h) &	89.46 (0.33h) &	83.74 (0.87h) &	86.07 (0.87h) &	89.66 (0.87h) & 30.4$\times$ \\ 
      & DFQ & 94.61 (8.79h) & 90.84 (1.50h) & 86.14 (0.75h) & 91.69 (0.75h) & 92.01 (0.75h) & 18.9$\times$\\ 
      & \bf Fast$_{\bf2}$ & 92.62 (0.06h) & 84.67 (0.03h) & 88.36 (0.03h) & 89.56 (0.03h) & 89.68 (0.03h) & 655.0$\times$ \\ 
      & \bf Fast$_{\bf5}$ & 93.63 (0.14h) & 89.94 (0.08h) & 88.90 (0.08h) & 92.04 (0.09h) & 91.96 (0.08h) & 247.1$\times$ \\ 
      & \bf Fast$_{\bf10}$ & 94.05 (0.28h) & 90.53 (0.15h) & 89.29 (0.15h) & 92.51 (0.17h) & 92.45 (0.17h) & 126.7$\times$  \\ 
      \hline
      
      \bf \multirow{11}{*}{\rotatebox[origin=c]{90}{CIFAR-100}} & Teacher  & 78.05  & 71.32  & 75.83  & 75.83    & 75.83 & - \\ 
      & Student  & 77.10 & 77.10       & 65.31  & 72.19    & 73.56   & - \\ 
      & KD & 77.87  & 75.07  & 64.06  & 68.58  & 70.79  & - \\ 
      & DeepInv$_{2k}$ & 61.32 (42.1h)   & 54.13 (20.1h)  & 53.77 (17.0h)  & 61.33 (21.9h) & 61.34 (18.2h) & 1.0$\times$ \\
      & CMI$_{500}$ & 77.04 (19.2h) & 70.56 (11.6h) & 57.91 (13.3h) & 68.88 (14.2h) & 68.75 (13.9h) & 1.6$\times$ \\ 
      & DAFL  & 74.47 (2.73h)   & 54.16 (0.73h) & 20.88 (1.67h)  & 42.83 (1.80h) & 43.70 (1.73h) & 15.2$\times$ \\ 
      & ZSKT  & 67.74 (1.67h)   & 54.31 (0.33h)  & 36.66 (0.87h) & 53.60 (0.87h)  & 54.59 (0.87h) & 30.4$\times$ \\ 
      & DFQ  & 77.01 (8.79h)   & 66.21 (1.54h)  & 51.27 (0.75h) & 54.43 (0.75h)  & 64.79 (0.75h) & 18.8$\times$ \\ 
      & \bf Fast$_{\bf2}$ & 69.76 (0.06h) & 62.83 (0.03h) & 41.77 (0.03h) & 53.15 (0.04h) & 57.08 (0.04h) & 588.2$\times$ \\ 
      & \bf Fast$_{\bf5}$ & 72.82 (0.14h) & 65.28 (0.08h) & 52.90 (0.07h) & 61.80 (0.09h) & 63.83 (0.08h) & 253.1$\times$ \\ 
      & \bf Fast$_{\bf10}$ & 74.34 (0.27h) & 67.44 (0.16h) & 54.02 (0.16h) & 63.91 (0.17h) & 65.12 (0.17h) & 124.7$\times$ \\ 
      \hline
  \end{tabular} 
  \caption{Student accuracy (\%) on 32$\times$32 CIFAR. The methods ``Teacher'', ``Student'' and ``KD'' is conducted on original training data and do not require data synthesis. DAFL, ZSKT and DFQ train generative networks for synthesis, while DeepInv, CMI and Fast (ours) optimizes batches to craft different samples.}
  \label{tbl:benchmark_classification}
\end{table*}

\begin{table*}[t]
  \centering
  \small
  \begin{tabular}{l c c c c c c}
     \toprule
     \bf \multirow{2}{*}{Method} & \bf \multirow{2}{*}{Data Amount} & \bf \multirow{2}{*}{Syn. Time} & \bf \multirow{2}{*}{Speed Up} &  \bf ResNet-50 & \bf ResNet-50 & \bf ResNet-50 \\
     & & & & \bf ResNet-50 & \bf ResNet-18 & \bf MobileNetv2 \\
     \hline  
     Scratch & 1.3M & - & - & 75.45 & 68.45 & 70.01 \\
     Places365+KD & 1.8M &  - & - & 55.74  & 45.53 & 39.89  \\
     Generative DFD & - & $\sim$300h & 1$\times$ & 69.75  & 54.66  & 43.15 \\
     DeepInv$_{2k}$ & 140k & 166h & 1.8$\times$ & 68.00 & - & - \\
     \bf Fast$_{\bf 50}$ & 140k & 6.28h & 47.8$\times$ & 68.61  & 53.45 & 43.02  \\
     \hline
  \end{tabular} 
  \caption{Student accuracy (\%) on 224$\times$224 ImageNet. We use an off-the-shelf ResNet-50 as the teacher model and train student models from scratch following the training protocol of \cite{yin2019dreaming}.} \label{tbl:imagenet}
\end{table*}

\subsection{Results on Classification}

\paragraph{CIFAR-10 \& CIFAR-100.} The student accuracy obtained on CIFAR-10 and CIFAR-100 datasets are reported in Table \ref{tbl:benchmark_classification}. In the table, baseline ``Teacher'', ``Student'' and ``KD'' train networks with the original training data, which does not require data synthesis. To verify the effectiveness of FastDFKD, we compare it with two types of data-free algorithms, including generative methods that train generators for synthesis and non-generative methods that optimize mini-batches iteratively. As shown in Table \ref{tbl:benchmark_classification}, generative methods are usually 10$\times$ faster than non-generative methods like DeepInv and CMI since they only need to train a single generator for synthesis. However, due to the limited capacity of the generative network, it is almost challenging to synthesize a diverse dataset for training, which may degrade the student performance. We find that the performance of generative methods tends to degrade as the complexity of the dataset increases from CIFAR-10 to CIFAR-100. By contrast, non-generative is usually more flexible than generative ones and thus more applicable to different tasks.  

Like non-generative methods, the proposed FastDFKD also optimizes mini-batches for data synthesis yet optimizes a generative network for adaptation. As shown in Table \ref{tbl:benchmark_classification}, the 5-step FastDFKD, i.e., Fast$_5$, can achieve 10$\times$ acceleration compared to existing generative methods and even more than $100\times$ acceleration compared to non-generative methods. For example, DeepInv$_{2k}$ synthesizes images by optimizing mini-batches, each of which requires 2,000 iterations to converge~\cite{yin2019dreaming}. To obtain 50,000 training samples for CIFAR, DeepInv$_{2k}$ would take 42.1 hours for data synthesis on a single GPU. by contrast, our method, i.e., Fast-5, adopts the same inversion loss as DeepInv but only requires 5 steps for each batch owning to the proposed common feature reusing, which is much more efficient than Deep Inversion. In addition to the improvements in efficiency, FastDFKD also achieves comparable or even superior performance compared to the state-of-arts. 

\paragraph{ImageNet.} To verify the effectiveness of FastDFKD, we further evaluate our method on a more challenging dataset, i.e., ImageNet, which contains 1.3 million training images of $224\times224$ resolutions from 1,000 categories. ImageNet is obviously much more complicated than CIFAR and thus much more time-consuming for data-free training. As shown in Table \ref{tbl:imagenet}, we compare our methods with a data-driven KD that uses Places365 as alternative data, and two data-free methods, i.e., Generative DFD~\cite{luo2020large} and DeepInv~\cite{yin2019dreaming}. Notably, Generative DFD~\cite{luo2020large} trains one generator of 224$\times$224 resolutions for each category, which leads to 1,000 generators in total. Although each generator can be optimized within one hour, the whole training process for 1,000 generators is still cumbersome. By contrast, our method only requires 6.28 hours for image synthesis and preserves comparable performance to existing methods. 

\begin{figure*}[t]
  \begin{center}
    \includegraphics[width=0.9\linewidth]{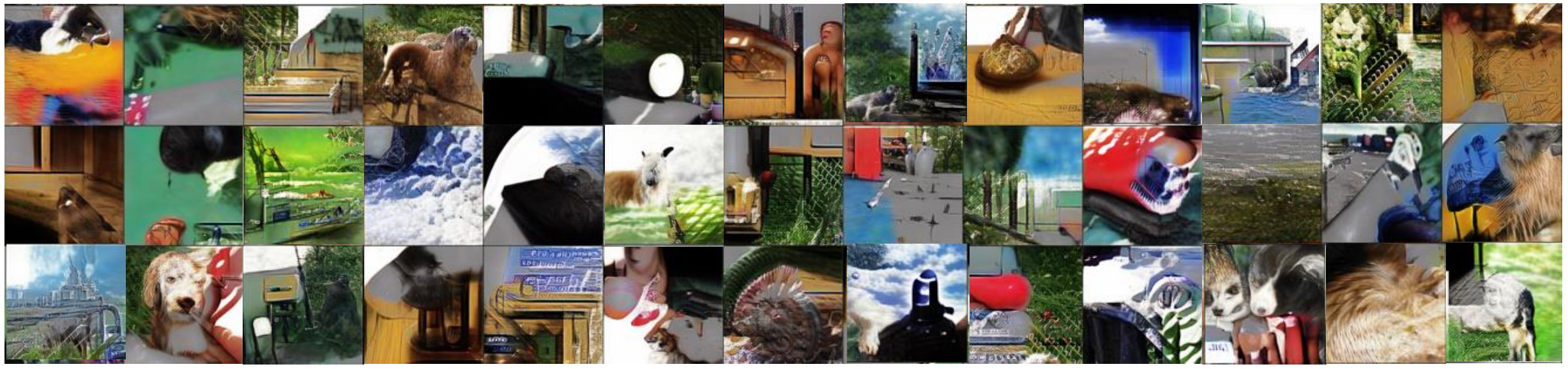}
  \end{center}
     \caption{Visualization of synthetic data, inverted from an off-the-shelf ResNet-50 classifier pre-trained on ImageNet. All samples are obtained using the 50-step FastDFKD.} \label{fig:imagenet_vis} \label{fig:tsne}
\end{figure*}

\subsection{Results on Segmentation} In this work, we further conduct data-free training on semantic segmentation tasks to show the effectiveness of our method. In segmentation, we only use the feature regularization loss and adversarial loss of \ref{eqn:deepinv} for data synthesis. The mIoU of the student model, as well as the data amount and synthesis time, are reported in Table \ref{tbl:newv2_seg}. We compare our method with DFAD~\cite{fang2019data}, DAFL~\cite{chen2019data} and DFND~\cite{chen2021learning}.
DFND is a data-driven method and assumes that a sufficient unlabeled set is available for in-domain data retrieval~\cite{chen2021learning}. DFAD and DAFL refer to data-free methods that train generative networks for knowledge distillation. In comparison, our method successfully synthesizes a training set only in 0.82 hours, which is much more efficient than DAFL (3.99 hours) and DFAD (6.0 hours).

\begin{table}[t]
  \centering
  \small
  \begin{tabular}{l l c c}
      \toprule
      \bf Method  & \bf Data Amount & \bf Syn. Time & \bf mIoU \\
      \hline  
      Teacher & 1,449 (NYUv2) & - & 0.519 \\
      Student & 1,449 (NYUv2) & - & 0.375 \\
      KD     & 1,449 (NYUv2) & - & 0.380 \\
      DFND  & 14 M (ImageNet) & - & 0.378 \\
      \hline
      DFAD    & 960k (GAN) & 6.0h & 0.364 \\
      DAFL    & 960k (GAN) & 3.99h & 0.105  \\
      \bf Fast$_{\bf 10}$ & 17K (Synthetic) & 0.82h & 0.366\\
      \hline
    \end{tabular} 
  \caption{Mean IoU the student model trained on NYUv2 Segmentation dataset. The teacher model is a Deeplabv3-ResNet50 network with ImageNet pre-training, and the student model is a freshly initialized Deeplabv3-Mobilenetv2.}\label{tbl:newv2_seg}
\end{table}

\begin{table}[t]
  \centering
  \small
  \begin{tabular}{l c c c}
     \toprule
     \bf Steps & \bf 2 steps & \bf 5 steps & \bf 10 steps  \\
     \hline  
     Teacher & 75.83 & 75.83 & 75.83 \\
     Student & 65.31 & 65.31 & 65.31\\
     DeepInv & 2.61 (0.03h) & 4.84 (0.06h) & 6.77 (0.11h)  \\
     CMI & 14.62 (0.05h) & 20.08 (0.12h) & 32.43 (0.22h)   \\
     \bf Fast & 41.77 (0.03h) & 52.90 (0.07h) & 54.02 (0.16h)  \\
     \hline
  \end{tabular} 
  \caption{Few-step experiments on CIFAR-100.} \label{tbl:few-step}
\end{table}

\begin{table}[t]
  \centering
  \small
  \begin{tabular}{l c c c}
     \toprule
     \bf Settings & \bf CIFAR-10  & \bf CIFAR-100 &   \\
     \hline  
     No Reuse + Pixel  & 44.74 & 3.11  \\
     No Reuse + GAN   & 87.33 & 35.48  \\
     Seq. + Pixel    & 71.97 & 10.93  \\
     Seq. + GAN   & 90.91 & 59.38   \\
     Meta + GAN   & 91.79 & 61.43   \\
     Meta + GAN + MMT  & 91.96 & 63.83 \\
     \hline
  \end{tabular} 
  \caption{Ablation study for FastDFKD, where MMT means deep inversion with momentum feature regularization.} \label{tbl:ablation}
\end{table}

\subsection{Quantitative Analysis}
\paragraph{Few-step synthesis.} As aforementioned, FastDFKD allows efficient data synthesis within only a few steps. This experiment validates our method by comparing it to the ``few-step'' versions of existing non-generative methods. For example, we reduce the optimization steps of original DeepInv~\cite{yin2019dreaming} from 2,000 to $\{10, 5, 2\}$ for CIFAR, which leads to the ``efficient'' DeepInv.
As shown in Table \ref{tbl:few-step}, the student accuracy of DeepInv and CMI severely degrades when the optimization steps are reduced, which means that existing methods fail to complete the data synthesis in only a few steps. By contrast, the proposed FastDFKD works well even when a 2-step optimization is deployed, which provides strong evidence for the effectiveness of FastDFKD.

\paragraph{Ablation Study.} In Table \ref{tbl:ablation}, we make an ablation study to make a systematic exploration for the proposed method, where three reusing strategies in Figure \ref{fig:framework} are considered: 1) no feature reuse; 2)  sequential feature reuse; 3) the proposed common feature reuse. Further, the effectiveness of the generative network is also verified by replacing it with the pixel updating proposed in \cite{yin2019dreaming}.

\paragraph{Visualization.} The synthetic results on ImageNet are visualized in Figure \ref{fig:imagenet_vis}, where all samples are obtained by deploying the 50-step FastDFKD on an off-the-shelf ResNet-50 classifier. Compared with existing methods that either require solving a 2000-step mini-batch optimization~\cite{yin2019dreaming} or train 1000 generative models for synthesis~\cite{luo2020large}, our proposed method can craft plausible samples within a few steps.

\section{Conclusions}
In this work, we propose a novel approach, termed as FastDFKD, to learn a meta-generator for fast data-free knowledge distillation, which is able to achieve 10 $\times$ and even 100$\times$ acceleration on CIFAR, NYUv2, and ImageNet. As the first attempt to improve the efficiency of data-free learning, the proposed approach successfully crafted a synthetic ImageNet in 6.28 hours, making data-free KD more applicable in real-world applications.

\paragraph{Acknowledgements.}
This work is supported by National Natural Science Foundation of China (U20B2066, 61976186,), Key Research and Development Program of Zhejiang Province (2020C01023), the Major Scientific Research Project of Zhejiang Lab (No. 2019KD0AC01), the Fundamental Research Funds for the Central Universities, Alibaba-Zhejiang University Joint Research Institute of Frontier Technologies,
NUS ARTIC~(Project Reference: ECT-RP2),
and Faculty Research Committee  Grant~(R-263-000-E95-133).

\clearpage
{
 \bibliography{citation}
}

\end{document}